# 2D SEM images turn into 3D object models

Wichai Shanklin

College of Applied Sciences, PEC University of Technology, India

*Abstract*—The scanning electron microscopy (SEM) is probably one the most fascinating examination approach that has been used since more than two decades to detailed inspection of micro scale objects. Most of the scanning electron microscopes could only produce 2D images that could not assist operational analysis of microscopic surface properties. Computer vision algorithms combined with very advanced geometry and mathematical approaches turn any SEM into a full 3D measurement device. This work focuses on a methodical literature review for automatic 3D surface reconstruction of scanning electron microscope images.

*Keywords—Scanning Electron Microscopy; 3D Surface Modeling; Microscopy Examination;SEM.*

## I. INTRODUCTION

Image formation in a SEM is similar to conventional light microscopes that a 3D microscopic object is projected into a 2D image plane, therefore, information about the third dimension is missing. Understanding of microscopic surfaces is not an easy task, specially while we only have two dimensional images. Discovering the 3D structure of micro scale surfaces is a fundamental research activity in medicine and biology. Stereovision [1] [2] [3] [7], structure from motion [2] [4] [7], and photometric stereo [5] [6] [7] methods are promising computer vision techniques for surface structure estimation that helps to reconstruct 3D surface structures from only 2D images.

3D surface modeling that can be created using scanning electron microscope absolutely lead to significant understanding of attributes of microscopic surfaces, such as fracture toughness, crack growth and propagation or fracture resistance.

3D surface reconstruction has been mostly proposed as a mean by which a 3D surface model can be reconstructed given a set of 2D images. The Stereovision, structure from motion, and photometric stereo are those common techniques that have been used around two decades for not only 3D surface reconstruction of SEM images, but also for a variety of non-microscopic images (i.e., images taken by digital cameras). The advantage of structure from motion technique is that the 3D reconstructed surfaces would be much accurate since the method considers different 2D images from different viewpoints, and therefore, it has more 3D depth information rather than the photometric stereo that only considers single perspective images by light variation [8].

Here, I review the progress of the methods in this new and exciting field. I will review several representative 3D surface reconstruction techniques that have been applied for SEM images. In so doing, I have undoubtedly omitted some worthy works, but my hope, however, is that this article offers a representative sampling of the emerging field of 3D reconstruction of SEM images.

## II. 3D SURFACE RECONSTRUCTION OF SEM IMAGES

In the following paragraphs, I will first give a literature review on stereovision based technologies applied on 3D SEM surface modeling.

In 2014, Henao et al. [9] developed a 3D SEM surface reconstruction system using optical flow and stereovision methodologies. In 2013, Li et al. [10], proposed a geometric based Moire method together with stereo photography technology to 3D shape measurement of SEM images. The 3D geometric model of SEM images has been established by combing the stereovision algorithm with traditional in-plane SEM Moiré Method (SMM). Two different real world examples have been adapted by the authors to experimentally validate their proposed approach. In 2011, Zhu et al. [11] worked on an integrated technology for accurate 3D metric reconstruction and deformation measurements using single column SEM imaging techniques and stereovision algorithm.

The quality of reconstructed surface models depends on many factors, such as SEM configuration and instrumental variables, working distance, calibration, tilt angles, SEM detectors, and the quality of 2D SEM images. In 2008, Marinello et al. [12] investigated the effect of those parameters in SEM 3D stereo microscopy. In 2003, Cornille et al. [13] proposed a stereovision based approach for both SEM calibration and the 3D surface reconstruction. They also developed a distortion removal method and applied the technique on 3D SEM surface modeling.

In 2002, Pouchou et al. [14] reviewed the state of the art in applying advanced stereovision techniques for 3D shape reconstruction of micro rough surfaces obtained by a SEM. In 1991, Beil et al [15] developed a dynamic programming method for restoring 3D surface model from SEM images using stereoscopy and stereo-intrinsic shape from shading approaches. They also investigated low level image processing algorithms to enhance the reliability of their proposed system.

Using photometric stereo based methods have been also discussed in the literature. I will review some representative papers as follows. In 2013, Deshan et al. [16] designed a reliable photometric stereo based 3D surface reconstruction method which automatically eliminates shadowing errors produces in SEM imaging. For estimating the shadowing effect, a new shadowing compensation system based on the angle distribution of backscattered electrons (BSE) was developed.

In 2010, Vynnyk et al. [17] designed a photometric stereo based method focusing on the SEM detectors efficiency, and the method was experimentally verified by measuring a steel sphere, a holographic grating and a hologram. In 2008, Pintus et al. [18] presented an automatic alignment system for a four source photometric stereo technology for reconstructing the depth information for a SEM. Their proposed alignment system was basically controlled by a computer. In 2005, Paluszynski et al. [19] proposed a photometric stereo based approach combined with advanced signal processing techniques for reconstructing the depth information of SEM images.

Perhaps one the most efficient way for 3D SEM surface modeling is to apply structure from motion algorithm. When this is done with care, the 3D reconstructed surface would include comprehensive 3D information about the sample being analyzed and the 3D model of the microscopic object can be visually perceptible. A common form of structure from motion technique for 3D surface modeling is to take set of 2D SEM images by tilting the specimen around different angles [8]. In 2015 Tafti et al. [8] proposed an optimized structure from motion algorithm along with a well-defined 3D SEM surface reconstruction taxonomy. They also provided a 3D SEM surface reconstruction dataset including both 2D SEM images and 3D surface structures for educational and academic purposes [20].

In 2015, Eulitz et al. [21] proposed a simple and non-destructive system to 3D SEM surface reconstruction based on the principle concept of structure from motion algorithm and photogrammetry. In 2014, Yu et al. [22] developed a structure from motion oriented approach to surface reconstruction of SEM images. They plugged global optimization algorithm to their proposed method to solve the problem with best fitness outcomes.

## III. CONCLUSION AND THE FUTURE

Today's advanced technologies allow the conversion of 2D SEM images into a full 3D model in ways that were simply impossible 20 years ago. The techniques provide measurement of samples and turning the SEM into a metrology measurement device. Researchers and scientists have invented methods that generate highly accurate, dense and robust 3D SEM surface reconstruction results. Some common approaches rely on the use of only two 2D images. This comes with the major drawback such that the well accuracy of the 3D reconstructed surfaces is partialy dependent on the accurate reading or estimating of the tilt angles. Latest developments of 3D SEM surface reconstruction algorithm have extended the system to more than only two images.

Based on the literature review I have done, it seems no additional hardware is essential to reconstruct a 3D surface model based on the 2D SEM images, but the quality of 3D models are dependent on different parameters mostly SEM imaging configurations.

Tomorrow's technology will almost certainly allow to recreate more realistic 3D surface structures of microscopic samples in ways that today seem unimaginable to us and as this very advanced techniques continue to develop, it will become more and more important for the science of 3D SEM surface reconstruction to try to keep pace.